\documentclass[conference]{IEEEtran}
\IEEEoverridecommandlockouts

\usepackage{cite}
\usepackage{amsmath,amssymb,amsfonts}
\usepackage{algorithmic}
\usepackage{graphicx}
\usepackage{textcomp}
\usepackage{xcolor}
\usepackage{subfigure}
\usepackage[algoruled,resetcount,linesnumbered]{algorithm2e}
\usepackage{multirow}
\usepackage{tabularx}

\SetAlFnt{\small} 
\SetKwInput{KwInput}{Input}                
\SetKwInput{KwOutput}{Output}              

\def\BibTeX{{\rm B\kern-.05em{\sc i\kern-.025em b}\kern-.08em
    T\kern-.1667em\lower.7ex\hbox{E}\kern-.125emX}}

\begin{document}

\title{Dynamic Clustering for Personalized Federated Learning on Heterogeneous Edge Devices}

\author{
\IEEEauthorblockN{Heting Liu, Junzhe Huang, Fang He, Guohong Cao}
  \IEEEauthorblockA{Department of Computer Science and Engineering, \\the Pennsylvania State University, PA, USA\\ E-mails: hetingl51@gmail.com, jmh8504@psu.edu, wh27hf@gmail.com, gxc27@psu.edu}
}
\maketitle
\begin{abstract}
Federated Learning (FL) enables edge devices to collaboratively learn a global model, but it may not perform well when clients have high data heterogeneity.
In this paper, we propose a dynamic clustering algorithm for personalized federated learning system ({\em DC-PFL}) to address the problem of data heterogeneity. 
DC-PFL starts with all clients training a global model and gradually groups the clients into smaller clusters for model personalization based on their data similarities.
To address the challenge of estimating data	
heterogeneity without exposing raw data, we introduce a discrepancy metric called {\em model discrepancy}, which approximates data heterogeneity solely based on the model weights received by	the server. We demonstrate that model discrepancy is strongly and positively correlated with data heterogeneity and can serve as a reliable indicator of data heterogeneity. 
To determine when and how to change grouping structures, we propose an algorithm based on the rapid decrease period of the training loss curve. 
Moreover, we propose a layer-wise aggregation mechanism that aggregates the low-discrepancy layers at a lower frequency to reduce the amount of transmitted data and communication costs.
We conduct extensive experiments on various datasets to evaluate our proposed algorithm, and our results show that DC-PFL significantly reduces total training time and improves model accuracy compared to baselines.
\end{abstract}


\vspace{-0.1in}
\section{Introduction}
In recent years, there has been significant growth in intelligent applications based on deep neural networks, driven by the availability of large amounts of data collected from the Internet.  However, the increasing concern about privacy and trust risks associated with this kind of data collection cannot be ignored~\cite{li2020federated}. Data generated by user activities on edge devices is often sensitive and should not be sent to the cloud directly.  To address this issue, Federated Learning (FL) has emerged, which allows edge devices (i.e., clients) to learn deep learning models without sharing their raw data~\cite{bonawitz-mlsys2019}.

Existing FL techniques exploit the computation capability of distributed devices and allow their models to be trained over a larger dataset. Clients learn a shared global model based on the local model weights. However, the problem of high data heterogeneity, i.e., not identically and independently distributed (non-IID) data in different clients, poses a significant challenge~\cite{ghosh-nips2020}\footnote{We use ``heterogeneous data'' and ``non-IID data'' interchangeably.}. Specifically, when different clients have non-IID data, the learned global model does not adapt to each client's local data distribution, which may reduce the average accuracy over all clients.

To address the problem of data heterogeneity, researchers have explored personalized FL (PFL) to learn personalized models for different clients to improve the overall accuracy. One approach to PFL is to conduct an additional fine-tuning step for each client after training a non-personalized global model \cite{arivazhagan-arxiv2019, fallah-nips2020}. However, the average accuracy of these personalized models is still largely influenced by the accuracy of the global model \cite{li-mobicom2021}. 
Another approach for FL personalization is through federated multi-task learning, which optimizes multiple models simultaneously \cite{smith-nips2017}. However, this approach has limited generalizability and only works well for certain types of tasks, such as training a support vector machine \cite{ouyang-tsn2022}.


To address the problems in PFL, one idea is to cluster clients with similar local data distributions into a group and utilize the data within the group to train personalized models for the clients. Some existing works in this direction generate client groups based on various rules and then let the clients within the same group share the same global model \cite{long-2023jwww}.
However, fixed groups may not work well in practice due to the tradeoff between the amount of data and the data heterogeneity within each group. For instance, clustering clients into fewer groups may result in larger data heterogeneity, and the learned model shared by the group may not adapt well to each client. 
On the other hand, clustering clients into more groups leads to lower data heterogeneity but less training data, causing overfitting. 
Moreover, determining suitable grouping structures under different FL tasks poses a challenge since the influence of data amount and data heterogeneity varies, and there is no prior knowledge available before training.

In this paper, we propose a dynamic clustering algorithm for personalized federated learning system ({\em DC-PFL}) to improve the average model accuracy. 
DC-PFL starts with all clients training a global model and gradually groups the clients into smaller clusters for model personalization based on their data similarities.
In the early phase of training, larger data amounts in each group help the model learn more general features, especially those in the shallow layers of the neural network \cite{hendrycks2019using}. In the later phase, lower data heterogeneity in each group adapts the model to each client's data distribution, improving model accuracy. 

There are two challenges in designing the dynamic clustering algorithm: (i) estimating the data heterogeneity of clients without exposing their raw data to each other and the server, and (ii) determining when and how to change the grouping structure. To address the first challenge, we introduce a discrepancy metric called 
{\em model discrepancy}, which estimates the data heterogeneity only based on the model weights received by the server. We show that the model discrepancy is highly positively correlated with the data heterogeneity and can be used as a good indicator of data heterogeneity.
To address the second challenge, we propose an algorithm based on the {\em Rapid Decrease Period (RDP)}, which identifies the time period when training loss drops rapidly. Using the radius of curvature of the training loss curve, we can identify the RDP and perform a possible group split based on performance comparisons, allowing for online adaptation of group structures.


To optimize communication overhead, we also investigate the discrepancy of each layer and observe that the discrepancies of different layers vary greatly. Based on this observation, DC-PFL adopts a layer-wise aggregation mechanism that aggregates the low-discrepancy layers at a lower frequency. This reduces the amount of data transmitted in each round and leads to an overall reduction in communication costs.

This paper has the following main contributions.

\begin{itemize}

    \item We identify that clustering clients with similar data distributions helps address the data heterogeneity problem through experiments. More importantly, we raise the idea of adaptively changing grouping structure in the training process to improve the model accuracy.
    
    \item We propose DC-PFL, a dynamic clustering algorithm with layer-wise aggregation for PFL to both improve model accuracy and reduce communication overhead.
    
    \item We conduct extensive experiments on various datasets. DC-PFL significantly reduces total training time and improves model accuracy compared to baselines.

\end{itemize}

\section{Background and Motivation}
\label{sec:motivation_background}

\subsection{Personalized Federated Learning with Heterogeneous Data}
Assume there are $n$ clients and a central server in the FL system. Each client $i \in \{1, \cdots, n\}$ has its own dataset $D_i$ sampled from distributions $P_1$, respectively.
All $n$ clients share the same model structure $\mathcal{M}$ and the learned global model is denoted by $\mathcal{M}(\mathbf{w})$, where $\mathbf{w}$ represents model parameters. The accuracy achieved by $\mathcal{M}(\mathbf{w})$ on the test set of client $i$ as $A_i$.
The primary objective of FL is to collaboratively train $\mathcal{M}(\mathbf{w})$ to maximize the average accuracy of $A_1, \cdots, A_n$.
Instead of learning a global model, PFL aims to address the data heterogeneity in FL by
learning $n$ models of the same structure $\mathcal{M}$, but personalized to $n$ different sets of model parameters $\mathbf{w_1}, \cdots, \mathbf{w_n}$ for $n$ clients, respectively.   
For client $i$, $\mathcal{M}(\mathbf{w_i})$ denotes the personalized model, 
$A^*_i$ denotes the best accuracy of $\mathcal{M}$ on $P_i$ with optimal parameters $w^*_i$. 
The objective of PFL is to collaboratively train $\mathcal{M}(\mathbf{w_i})$ such that $A_1, \cdots, A_n$ are close to $A^*_1, \cdots, A^*_n$.

\begin{figure}[htbp]
    \centering
    \includegraphics[width=2.2in]{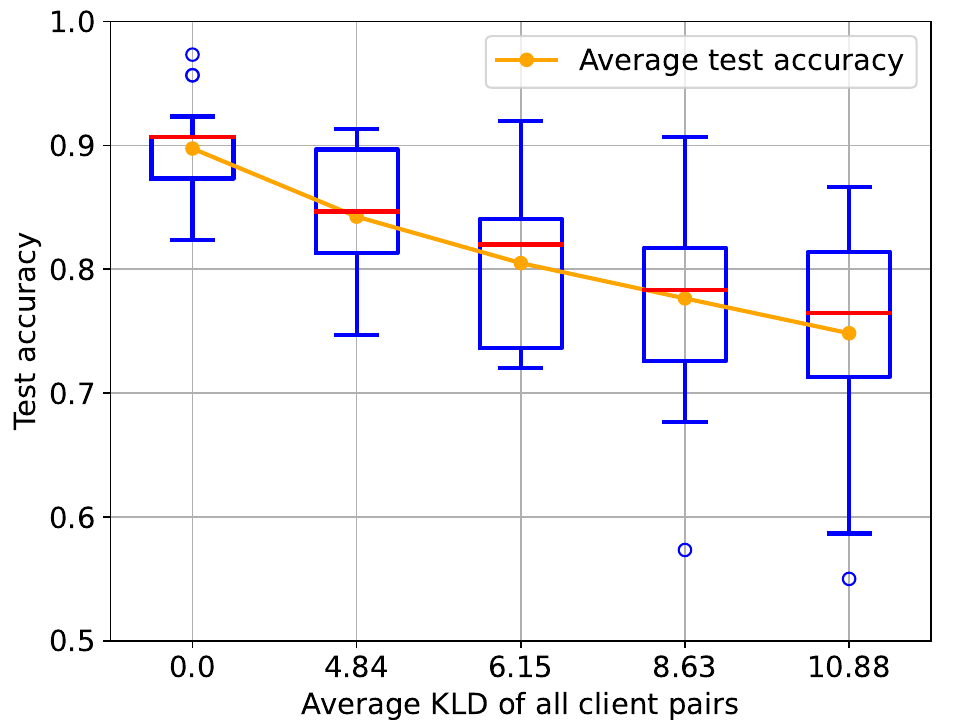}
    \vspace{-0.1in}
    \caption{Accuracy of the global models under different data heterogeneity.}
    \label{fig:dataKLD_accuracy}
    \vspace{-0.2in}
\end{figure}

\subsection{Impact of Data Heterogeneity on FL}
\label{sec:datahetero_fl}

In FL, the performance of the global model degrades with high data heterogeneity across clients \cite{hanzely-arxiv2020federated, fallah-nips2020, huang-aaai2021}.
To show the impact of data heterogeneity, we use FedAvg \cite{mcmahan-aistats2017} to train a series of global models on CIFAR-10 in different data heterogeneity settings. 
Specifically, we train each global model with 50 clients, and the datasets of different clients are sampled from different distributions. 
We measure the level of data heterogeneity among clients by computing the Kullback–Leibler divergence (KLD) \cite{huang-nips2016} between the distributions of each pair of clients and then averaging the KLD across all pairs of clients.
Formally, for discrete distributions $P_i$ and $P_j$ defined in the same probability space $\mathcal{X}$, the KLD from $P_i$ to $P_j$ is 
\begin{equation}\small
\setlength{\abovedisplayskip}{3pt}
\setlength{\belowdisplayskip}{3pt}
\label{eq:pair_kld}
    D_{KL}(P_i||P_j) = \sum_{x\in\mathcal{X}} P_i(x)\log(\frac{P_i(x)}{P_j(x)}).
\end{equation}
Since KLD is not symmetric, the data heterogeneity of client $i$ and client $j$ is defined as $D_{KL}(i, j) = D_{KL}(j, i) = \frac{1}{2}(D_{KL}(P_i||P_j) + D_{KL}(P_j||P_i))$. 
The data heterogeneity of all $n$ clients in a group is defined as the average of the data heterogeneity among all client pairs within the group, i.e.,

\begin{equation}\small
\setlength{\abovedisplayskip}{3pt}
\setlength{\belowdisplayskip}{3pt}
\label{eq:avg_kld}
    \bar{D}_{KL} = \frac{2}{n}\sum_{i=2}^{n}\sum_{j=1}^{i-1} D_{KL}(i,j).
\end{equation}

We run the experiments several times. Each time, we sample a dataset from a different distribution for each client, record $\bar{D}_{KL}$, and train the global model.
We calculate the average of the test accuracy on each client as the final model accuracy.
Fig.~\ref{fig:dataKLD_accuracy} plots data heterogeneity (i.e., the average KLD of all client pairs) and the corresponding accuracy achieved by the global model for all clients. 
As shown, the accuracy significantly decreases as the average KLD increases, indicating that high data heterogeneity negatively impacts FL performance.
We also notice that the variance of the accuracy achieved by different clients increases with the data heterogeneity, 
suggesting that it is difficult to train a common global model that performs well for all clients with heterogeneous data.

\vspace{-0.05in}
\subsection{Group-based Personalized FL}
\label{sec:groupbased_fl}

To mitigate the negative impact of data heterogeneity, we propose a simple group-based approach for PFL, which clusters clients with similar data distributions into the same group, facilitating the training of better personalized models for the group.
Several clustering algorithms exist for clustering a set of subjects with a pre-defined measurement of dissimilarity between two subjects, such as K-means \cite{hartigan-kmeans1979}, spectral clustering \cite{dhillon-kdd2004}, and hierarchical clustering \cite{murtagh-Wiley2012}. 
Among them, both K-means and spectral clustering require the number of clusters to be pre-known, which is not suitable for clustering clients with arbitrary data distributions. 
Hence, we propose using hierarchical clustering of agglomerative nesting controlled by a single parameter, the distance threshold (dissimilarity) $\gamma$. It iteratively merges two most similar groups into a new group when the distance between them does not exceed $\gamma$.

Specifically, we cluster the clients into several groups based on hierarchical clustering as follows:
\textbf{1) Initialization}: Each client forms a group by itself, resulting in $n$ groups initially; 
\textbf{2) Merge clusters}: We select the two groups with the smallest distance and merge them to form a new group. The distance of two groups is defined as the average of all pairwise dataset KLD between the clients in the two groups (each client in a pair comes from a different group); 
and \textbf{3) Termination}: The algorithm terminates if the smallest distance between any two groups exceeds $\gamma$, and repeats step 2 otherwise.

After clustering, Clients within the same group share a model trained with the data in the group. The groups are fixed during the model training process. 
To investigate the impact of $\gamma$ on the performance of group-based PFL, we vary its setting and observe the final model accuracy~\footnote{Throughout this paper, ``accuracy'' refers to accuracy on the test set unless otherwise specified.}. 

\begin{figure}[t]
    \begin{minipage}[t]{0.23\textwidth}
    \centering
    \includegraphics[width=1.1\textwidth]{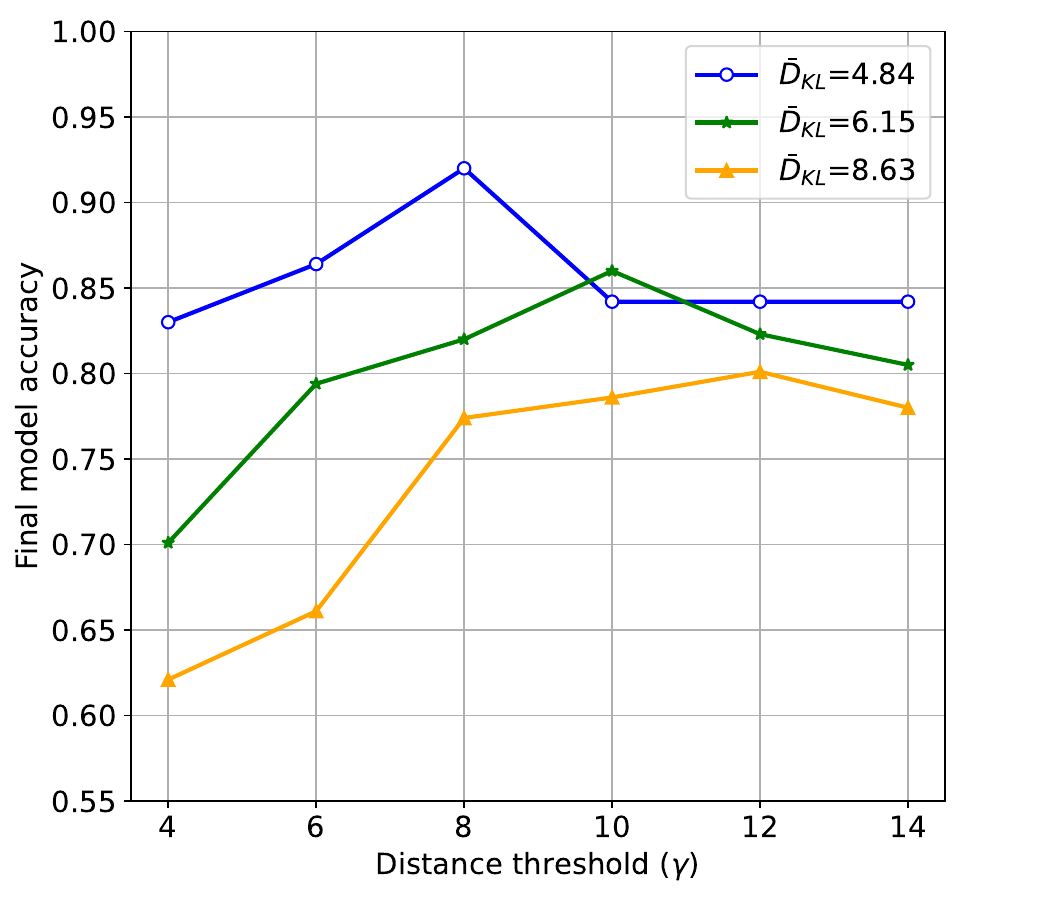}
    \vspace{-0.2in}
    \caption{Accuracy under \\different $\gamma$.}
    \label{fig:group_gamma_accuracy}
    \end{minipage}\; \begin{minipage}[t]{0.23\textwidth}
    \centering
    \includegraphics[width=1\textwidth]{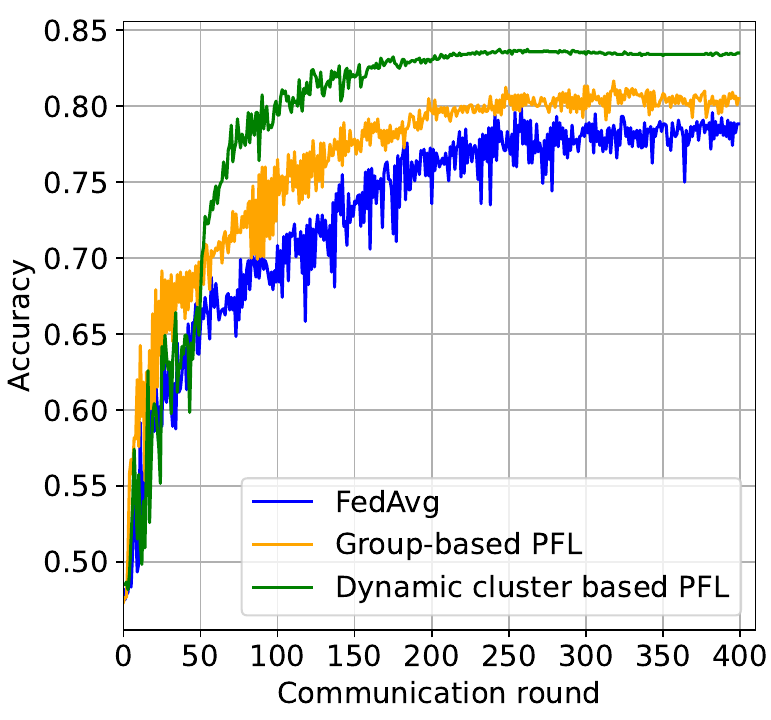}
    \vspace{-0.2in}
    \caption{Accuracy achieved by different FL algorithms.}
    \label{fig:dynamic_group_accuracy}
    \end{minipage}
    \vspace{-0.2in}
\end{figure}

We use the same client settings in Section \ref{sec:datahetero_fl}. Fig.~\ref{fig:group_gamma_accuracy} shows the final model accuracy under different levels of data heterogeneity and $\gamma$. 
We have three observations: 
\textbf{i)} Group-based PFL outperforms FedAvg when $\gamma$ is properly set, indicating the potential of group-based PFL to improve final model accuracy. When $\gamma$ is large enough, all clients are clustered into one group, and only one global model is trained for all clients, which is equivalent to FedAvg. When $\bar{D}_{KL}$ is 4.84, all clients are clustered into one group when $\gamma$ is greater than 10. We observe that with $\gamma$ set to 8, the final model accuracy is 92.27, much better than 84.35 with $\gamma$ set to 10. We have similar observations when $\bar{D}_{KL}$ is 6.15 or 8.63.
\textbf{ii)} We observe that the $\gamma$ that achieves the best accuracy varies under different levels of data heterogeneity. 
For example, when $\bar{D}_{KL}=4.84$, the best $\gamma$ is 8, while when $\bar{D}_{KL}=8.63$, the best $\gamma$ is 12. This observation suggests that we cannot find a single optimal $\gamma$ for all levels of data heterogeneity. Instead, we may set $\gamma$ based on the level of data heterogeneity to improve final model accuracy.
\textbf{iii)} Under high data heterogeneity (i.e., $\bar{D}_{KL} = 8.63$), group-based PFL only achieves limited accuracy improvement compared to FedAvg, no matter how $\gamma$ is set (i.e., 80.1\% v.s. 77.6\% when $\gamma$=12). 
The reason is that, with a large $\bar{D}_{KL}$ and a small $\gamma$, we may face severe overfitting issues (e.g., each group has only one client). However, if we pre-set a large $\gamma$ to ensure a large amount of data in each group, each group will have high data heterogeneity, and it may not be suitable to learn a shared model for all clients in the group. In short, when $\bar{D}_{KL}$ is large, a fixed $\gamma$ is not sufficient to ensure both low data heterogeneity and a large amount of train data in each group.


\vspace{-0.05in}
\subsection{Dynamic Clustering for Personalized FL (PFL)}
\label{sec:dynamic_cluster}

Group-based PFL is affected by two factors: the data heterogeneity of each group and the amount of training data in each group. The impacts of these two factors on PFL vary in different training rounds: 
i) in the early training rounds, the model needs more training data to learn general-purpose features in the shadow layers \cite{hendrycks2019using}; and ii) in the later training rounds, PFL personalizes models to adapt to different clients by training the models within a group of clients with lower data heterogeneity.
Therefore, in this paper, we explore a dynamic clustering scheme to adjust $\gamma$ in different training rounds to train models with more data in the early phases and with lower heterogeneous data in the later phases for PFL.

Before presenting our DC-PFL, we discuss a naive dynamic clustering scheme to demonstrate its ability to improve accuracy. In this scheme, all clients train a global model together at the first 50 training rounds (a pre-set parameter), i.e., the same manner as FedAvg. 
After that, clients are clustered into two groups ($\gamma$ set as 12) and continue to train models within each group. 
Fig.~\ref{fig:dynamic_group_accuracy} shows the average accuracy of all clients achieved by FedAvg, group-based PFL, and dynamic clustering based PFL. We use the same settings as in Section \ref{sec:datahetero_fl} with $\bar{D}_{KL}$ set to 8.63.
We observe that the dynamic clustering based PFL performs consistently with FedAvg at the first 50 training round; and it achieves the highest accuracy when training is converged. 
In particular, there is a rapid increase in accuracy of dynamic clustering-based PFL when clients are split into groups right after $50^{th}$ round.
This validates that model training with more data in the early rounds (instead of adapting 
to different clients) helps extract general-purpose features in the shadow layers to improve the final accuracy.
In summary, these observations motivate the design of a dynamic clustering algorithm for PFL.

\vspace{-0.2in}
\section{DC-PFL: Dynamic Clustering for PFL}
\label{sec:dynamic_cluter}



\vspace{-0.15in}
\subsection{Data Heterogeneity and Model Discrepancy}
\label{sec:model_discrepancy}
\vspace{-0.1in}
Typically, local raw data and its distribution of a client are private to the server and other clients, so we may not calculate the data heterogeneity with equation~(\ref{eq:pair_kld}) and equation~(\ref{eq:avg_kld}).
Therefore, we propose a new method to measure the data heterogeneity given the model parameters after local training. 
Although the existing Centered Kernel Alignment (CKA) \cite{cortes-2012jmlr} technique measures model similarity, it requires passing input data to the model to generate representations. The model similarities are then determined by comparing these output representations, which involves the multiplication of input data with all model weights. This computation-intensive process can lead to significant costs and might not meet our algorithm's requirements\footnote{Additionally, we expect the metric to easily evaluate the similarity of layer-wise model weights used in Section~\ref{sec:commu_optimization}. CKA may not be a good indicator in the case of layers since the similarities of the output representations do not necessarily indicate the similarities of the model weights.}.



In FL, each client performs stochastic gradient descent (SGD) locally in each communication round to train the model shared by a group of clients. 
Intuitively, two clients with similar data distributions may have their local trainings tailored to similar directions in the parameter space and obtain similar models, and vice versa.
Thus, the dissimilarity among the model weights of a group of clients should be a good indicator of the data heterogeneity, i.e., the dataset KLD, of the group.
Formally, we define the \textit{model discrepancy} as a dissimilarity metric between the model parameters of client $i$ and client $j$ at round $t$ as follows:

\begin{equation}\small
\setlength{\abovedisplayskip}{3pt}
\setlength{\belowdisplayskip}{3pt}
\label{eq:model_disc}
   D^t_{i, j} = \frac{ |S(\mathbf{w}^t_i) - S(\mathbf{w}^t_j)|}{dim(\mathbf{w}^t_i)},
\end{equation}
where $\mathbf{w}_i$ and $\mathbf{w}_j$ are the model weights of client $i$ and client $j$ at round $t$ 
respectively, $S(\cdot)$ 
is a scaling function to normalize tensors with arbitrary values to $[0, 1]$, and $|\cdot|$ denotes the $L_1$-norm.
In this paper, we adopt the commonly used linear scaling \cite{he-arxiv2016scale}, 
i.e., $S(w) = \frac{w - \min(w)}{\max(w) - \min(w)}$ , where $\min(w)$ is the smallest value in $w$ and $\max(w)$ is the largest value in $w$. Note that in equation~(\ref{eq:model_disc}), $dim(\cdot)$ denotes the number of parameters in $\mathbf{w}$ and $D^t_{i,j}$ quantifies the difference between the local model of client $i$ and that of client $j$ at round $t$.
However, different from dataset KLD which is a constant as long as the clients' datasets are not changed, model discrepancy may change over time since the model weights are updated in each round.
To fill this gap, we monitor the model discrepancy calculated in each training round, and observe that it becomes stable after a starting phase, i.e., the first $T_{start}$ training rounds.
Thus, to obtain a more accurate estimation of $D_{KL}(i,j)$, we compute the average model discrepancy within the first $T_{start}$ rounds, denoted as $D_{i, j}$.

\begin{figure}[t]
    \vspace{-0.1in}
    \subfigure[Dataset KLD]{
    \centering
        \includegraphics[width=1.6in]{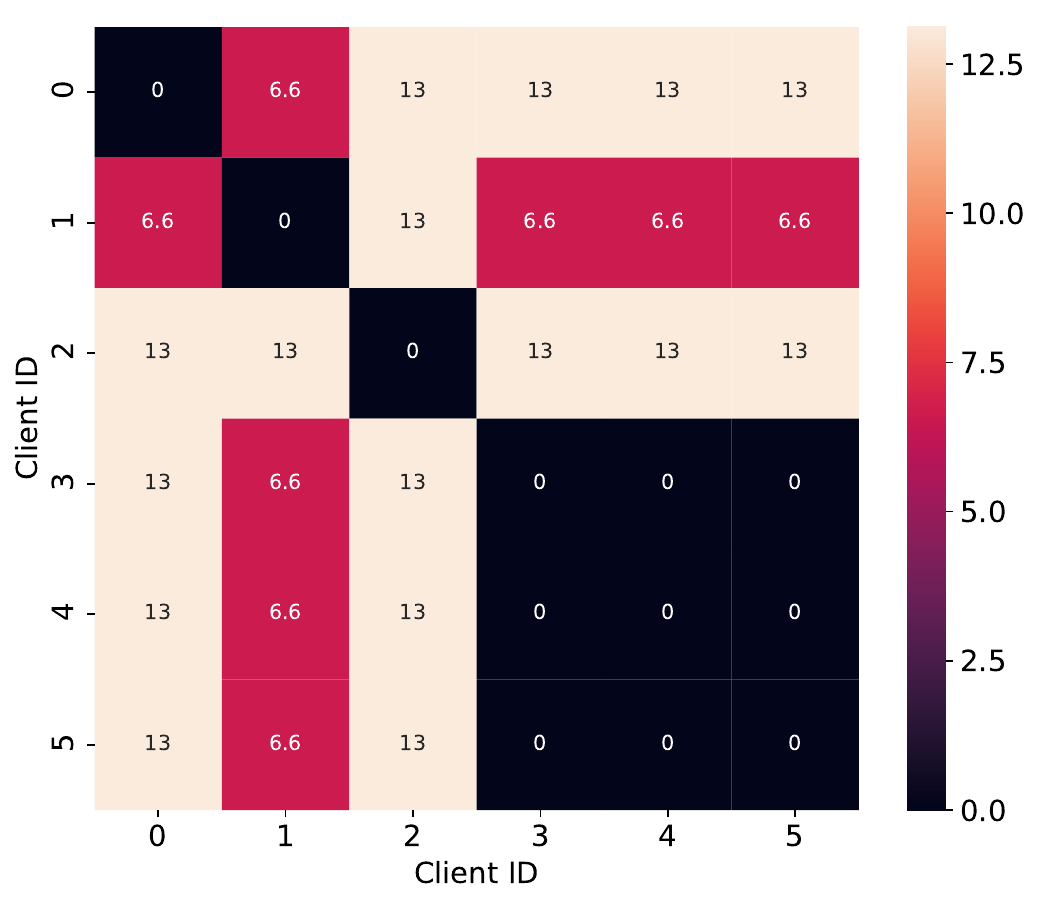}
    }
    \subfigure[Model discrepancy]{
    \centering
        \includegraphics[width=1.6in]{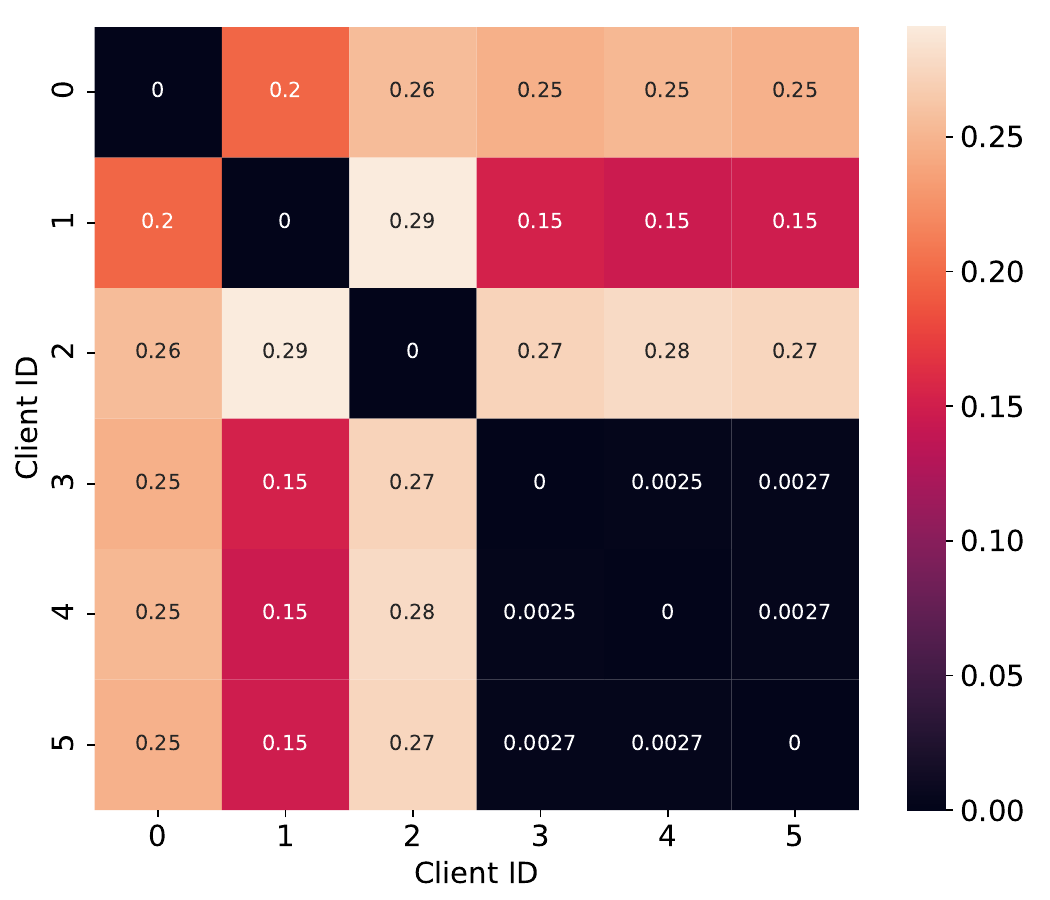}
    }
    \vspace{-0.1in}
    \caption{An example of the pair-wise dataset KLD and model discrepancy of six clients.}
    \label{fig:dataKLD_modelDisc}
    \vspace{-0.2in}
\end{figure}

\begin{figure}[t]
    \vspace{-0.1in}
    \subfigure[KLD-model discrepancy]{
    \centering
        \includegraphics[width=1.6in]{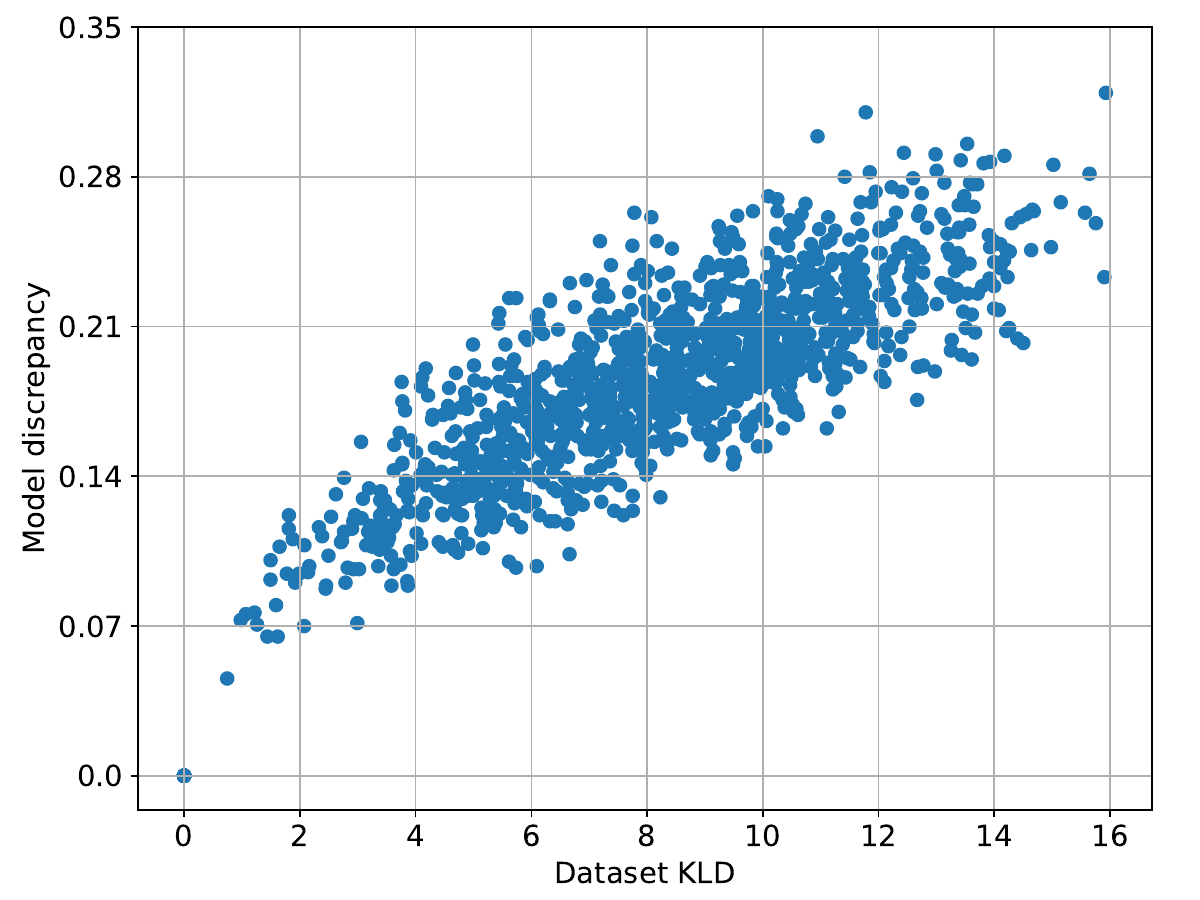}
    }
    \hspace*{-1.0em}
    \subfigure[Hierarchical group graph]{
    \centering
        \includegraphics[width=1.7in]{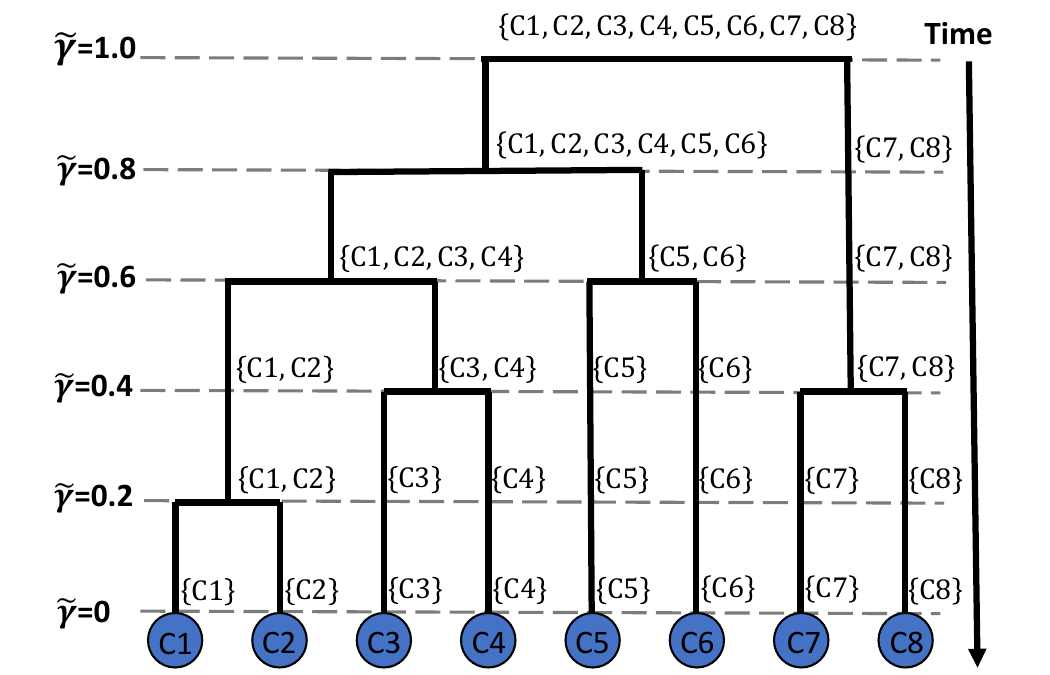}
    }
    \vspace{-0.1in}
    \caption{Illustrations of (a) Scatter plot of dataset KLD-model discrepancy pair, and (b) the hierarchical group graph.}
    \label{fig:hier_group_graph}
    \vspace{-0.2in}
\end{figure}

To verify the effectiveness of using the stable model discrepancy to estimate the data heterogeneity, we train an FL model with the same experiment setting in Section~\ref{sec:motivation_background}. For every pair of client $i$ and client $j$, we compute both their dataset KLD (i.e., $D_{KL}(i,j)$) and model discrepancy (i.e., $D_{i, j}$).
Fig.~\ref{fig:dataKLD_modelDisc} plots the matrices of pairwise dataset KLD and model discrepancy between six clients, respectively. We observe that the two matrices are similar, e.g., client 3 and client 4 have the most similar data distributions with $D_{KL}(i,j) = 0$, and they also have the smallest $D_{i,j}$ of 0.0025.
Furthermore, we make a more general observation based on data from 30 clients. Fig.~\ref{fig:hier_group_graph}(a) plots the relationship between the pairwise datasets KLD and model discrepancy. 
As can be seen, there is a clear positive correlation between $D_{i, j}$ and $D_{KL}(i,j)$, 
with a correlation coefficient of 0.895, suggesting that $D_{i, j}$ is a good indicator of $D_{KL}(i,j)$.


\vspace{-0.05in}
\subsection{Hierarchical Group Graph}
\label{sec:hierarchical_group_graph}


With the model discrepancy, the distance between two clusters of clients can be computed as the average of all pairwise model discrepancies of clients in the two clusters. 
With the distance computed, we 
follow the hierarchical clustering algorithm to build a hierarchical group graph in the same way introduced in Section~\ref{sec:groupbased_fl}.
Note that the structure of groups changes in accordance with the distance threshold $\gamma$. 
To control the dynamic clustering process, we first find the minimum $\gamma$ that clusters all the clients into the same group, denoted as $\gamma^{global}$. 
Then we build a mapping from a raw distance threshold $\gamma\in [0, \gamma^{global}]$ to a normalized distance threshold $\tilde{\gamma} \in [0,1]$ by $\tilde{\gamma} = \frac{\gamma}{\gamma^{global}}$.
As such,
we can obtain different structures of clustering, which form a hierarchical group graph shown in Fig.~\ref{fig:hier_group_graph}(b), by decreasing the value of $\tilde{\gamma}$ from 1 to 0. 
As shown, when $\tilde{\gamma} = 1$, all the clients are in one group. When $\tilde{\gamma}$ decreases, the clients are divided into more and smaller groups. And when $\tilde{\gamma}$ is decreased to a value small enough, each client forms one group. 
It is worth noting that if two clients are in the same group with a small $\tilde{\gamma}$, they are in the same group under any larger $\Tilde{\gamma}$.
Finally, the constructed hierarchical group graph is used as an input to the dynamic clustering algorithm (introduced later), i.e., given a $\tilde{\gamma}$, the corresponding groups are provided to the algorithm.

\vspace{-0.05in}
\subsection{The Dynamic Clustering Algorithm}
\label{sec:dcluster_algorithm}

The main challenge of designing the dynamic clustering algorithm lies in when and how to change the group structure in the training process. 
It is difficult because we cannot obtain the knowledge of the training task in advance and thus can only adapt the group structure in an online manner based on the knowledge revealed in the training process.

As described in Section \ref{sec:groupbased_fl}, the final accuracy can be improved by using more data in the early training phase (i.e., with a larger $\Tilde{\gamma}$) to capture more general features and using data of low heterogeneity (i.e., with a smaller $\Tilde{\gamma}$) for model personalization. 
Therefore, we propose an algorithm to dynamically adjust $\tilde{\gamma}$, 
which first trains the FL model with all clients and then gradually divides the clients into smaller groups 
in later training rounds.

\begin{figure}[t]
    \vspace{-0.1in}
    \subfigure[Training loss v.s. time during FL training (training loss curve)]{
    \centering
        \includegraphics[width=1.6in]{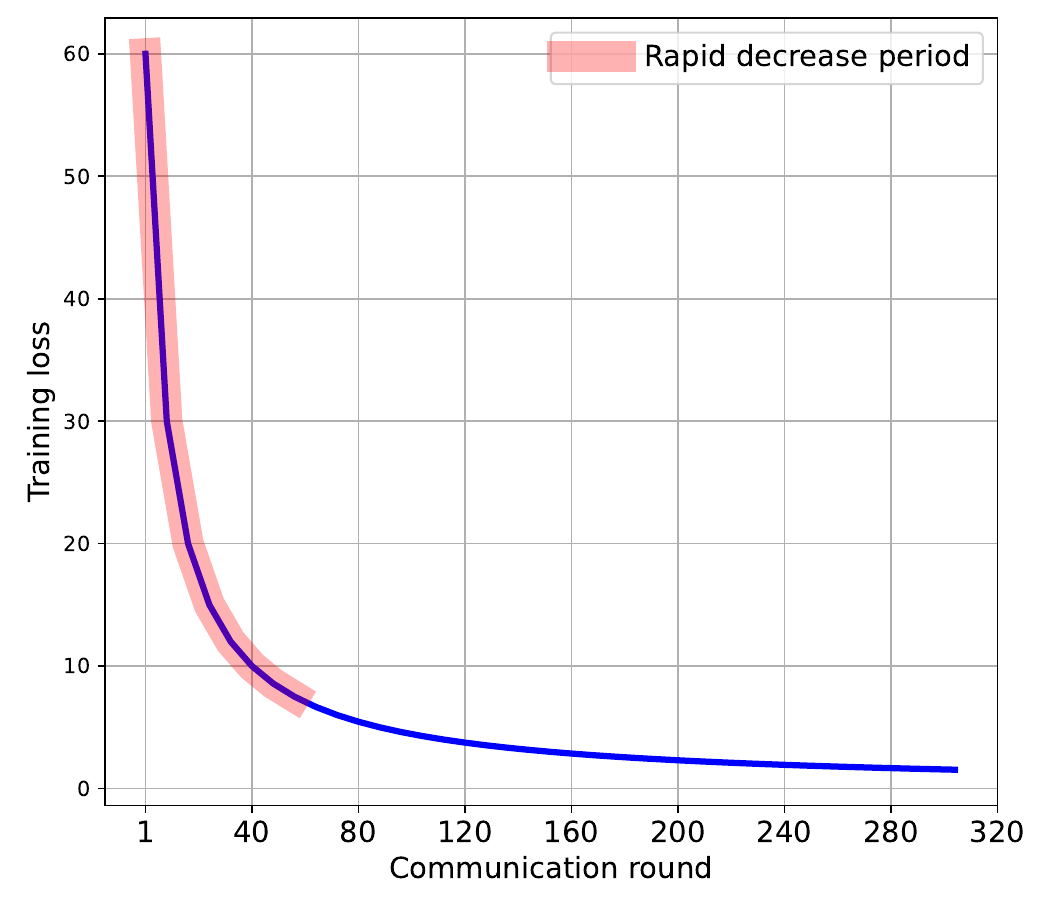}
    }
    \subfigure[Radius of curvature at each round.]{
    \centering
        \includegraphics[width=1.6in]{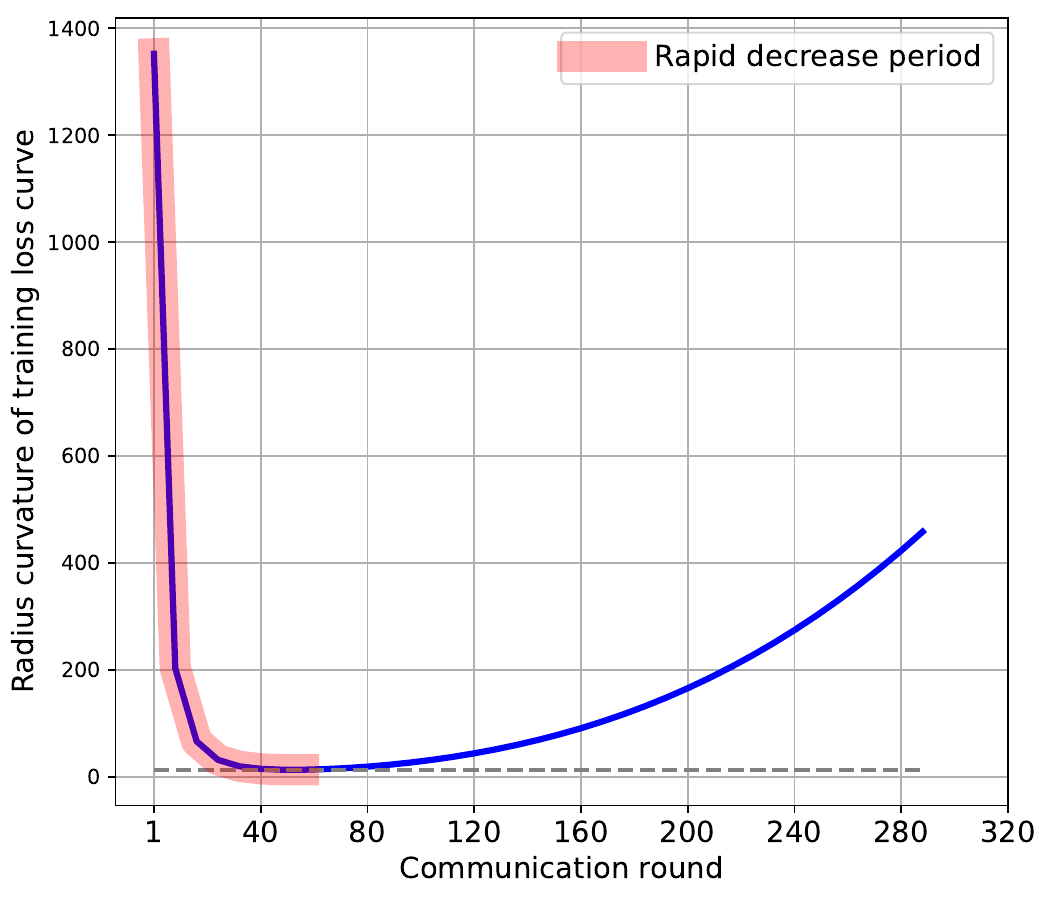}
    }
    \vspace{-0.1in}
    \caption{Illustration of the RDP.}
    \label{fig:rapid_decrease_period}
    \vspace{-0.3in}
\end{figure}

To decide when to split the group, we propose to identify the periods where the training loss decreases more rapidly, called {\em Rapid Decrease Periods (RDP)}, and split the groups according to the hierarchical group graph at the end of each rapid decrease period.
As shown in Fig.~\ref{fig:rapid_decrease_period}(a),
when using a fixed $\tilde{\gamma}$ for FL training, the training loss $l(t)$ 
typically first decreases quickly, gradually decreases slowly, and finally becomes stable. 
The RDP is defined as the period between the beginning of the training and the time point when the {\em radius of curvature} of the training loss curve is the minimum.
To obtain the radius of curvature, we calculate the first derivative and second derivative of the training loss curve, which are denoted as $l^{\prime}(t)$ and $l^{\prime\prime}(t)$, respectively. 
Then the radius of curvature $r(t)$ at round $t$ is defined as \cite{courant-radius2011}:
\begin{equation}\small
\setlength{\abovedisplayskip}{3pt}
\setlength{\belowdisplayskip}{3pt}
\label{eq:r_curvature}
    r(t) = \frac{(1+(l^{\prime}(t))^2)^\frac{3}{2}}{l^{\prime\prime}(t)}.
\end{equation}
We show the curve of $r(t)$ in Fig.~\ref{fig:rapid_decrease_period}(b). The red part represents the RDP, i.e., the period before the minimum $r(t)$.

In practice, calculating $r(t)$ at each round faces many challenges because 
the observations of training loss are more likely discrete points instead of a smooth curve shown in Fig.~\ref{fig:rapid_decrease_period}(a). 
We propose the following method to estimate the radius of curvature.
In the training round $t$, after client $i$ downloads the model from the group it belongs to in that round, it first tests the loss $l_i(t)$ on its dataset.
After the client sends $l_i(t)$ to the server, the server computes the average training loss of all clients following $l(t) = \frac{1}{n} \sum_{i=1}^n l_i(t)$.   
In order to reduce the randomness of $l(t)$ and make the observed training loss curve more smooth, we set a sliding window of length $s$, and calculate the average of $l(t)$ within the sliding window following $l_w(t) = \frac{1}{s}\sum_{t' = max(1, t-s)}^{t} l(t')$ as the smoothed loss at round $t$.
The sliding window moves forward by one communication round at each round. 
In our experiments, we set $s$ to be 5 communication rounds.

Next, we estimate the first derivative and second derivative of the training loss curve.
Since the physical meaning of the first derivative $l^{\prime}(t)$ is the change rate of $l(t)$, we compute the decrease rate of the loss to estimate $l^{\prime}(t)$. 
To achieve this goal, we measure the decrease rate of $l_w(t)$ by subtracting the average losses of two consecutive sliding windows, and $l^{\prime}(t)$ is estimated by $l^{\prime}(t) = l_w(t) - l_w(t-1)$. 
Similarly, the second derivative $l^{\prime\prime}(t)$ is estimated by $l^{\prime\prime}(t) = l^{\prime}(t) - l^{\prime}(t-1)$.
With $l^{\prime}(t)$ and $l^{\prime\prime}(t)$ obtained, the radius of curvature at round $t$ (i.e., $r(t)$) is computed following Eq.~(\ref{eq:r_curvature}).

\textbf{Identifying RDP.} Based on the radius of curvature, we identify RDP as follows. 
During the FL training process, $r(t)$ is computed every round at the server after the $s^{th}$ round. 
When the server detects that $r(t)$ is not decreasing for consecutive $t^{obv}$ rounds, i.e., $r(t) < r(t^\prime)$ for $\forall t^\prime = t+1, \cdots, t+t^{obv}$, it identifies that the current RDP ends\footnote{
Note that we only perform a constant number of operations in each
round which takes only a few milliseconds to identify RDP. It is negligible
compared to the computation time (seconds level).}.
Then, the server modifies $\tilde{\gamma}$ by adding a step parameter $\lambda_{\tilde{\gamma}}$, and splits the clients into groups following the results of hierarchical clustering obtained in Section~\ref{sec:hierarchical_group_graph} with the new $\tilde{\gamma}$.

With the new $\tilde{\gamma}$, the new group structure may not always be better than the old group structure, which may result in an increase of the training loss. To avoid the potential degradation brought by re-clustering, 
we compare the average loss of the new group structure and the old one to decide whether to evolve the group structure into the new one.
Specifically, suppose the server detects the end of the RDP at round $t_0$, and the current group structure is denoted $G^{0}$ under $\tilde{\gamma}_0$. 
In the next round of training, the server needs to decide whether to evolve the group structure into $G^{1}$ under $\tilde{\gamma}_1$. 
To achieve this, the server aggregates model weights for each group based on $G^{0}$ and $G^{1}$ to produce the model of each group (called group model) to send to the clients in the group. We denote the models received by client $i$ as $M^{0}_i$ and $M^{1}_i$, for client $i = 1, \cdots, n$, respectively. 
At the beginning of round $t_0+1$, every client downloads both group models $M^{0}_i$ and $M^{1}_i$, followed by a local SGD. 
Then it tests the training loss of $M^{0}_i$ and $M^{1}_i$, denoted by $l_i^{M^0}(t_0+1)$ and $l_i^{M^1}(t_0+1)$, respectively, and sends the computed losses to the server. 
The server computes the average losses of all clients corresponding to $G^{0}$ and $G^{1}$, which are 
$l^{M^0}(t_0+1) = \frac{1}{n} \sum_{i=1}^n l_i^{M^0}(t_0+1)$ 
and $l^{M^1}(t_0+1) = \frac{1}{n} \sum_{i=1}^n l_i^{M^1}(t_0+1)$.
If $l^{M^0}(t_0+1) > l^{M^1}(t_0+1)$, i.e., the loss decreases with the new group structure, 
the group structure will evolve into $G^1$ and each client $i$ uploads the model weights obtained based on $M^{1}_i$.
Otherwise, the old group structure $G^{0}$ is used for the next $t^{sp}$ training rounds. 

Whenever the group structure is changed, a new RDP is monitored and determined following the above strategy. All the steps are repeated until the training process ends, i.e., each client itself is a group and the model accuracy converges.

\vspace{-0.05in}
\section{Communication Optimization: Layer-Wise Aggregation}
\label{sec:commu_optimization}

In FL, a significant number of communication rounds between clients and the server are required, with millions of model weights transmitted in each round \cite{liu-infocom2023, Tu-sensys2021}
To address this, we propose a layer-wise discrepancy-based aggregation method for DC-PFL to reduce communication overhead.

As discussed in Section~\ref{sec:model_discrepancy}, model discrepancy $D^{t}_{i,j}$ is computed by the server at the beginning of training based on the received model weights from clients. However, the discrepancies between different layers in the model can vary significantly, which affects the frequency requirements for the aggregation of different layers. Therefore, we propose the concept of ``layer-wise discrepancy'' for group $g$ as follows:
\begin{equation}\small
\setlength{\abovedisplayskip}{3pt}
\setlength{\belowdisplayskip}{3pt}
    D^{t}_{g,l} = \frac{1}{m}\sum_{i=1}^{m} \frac{|S(\mathbf{w}_{l,i}) - S(\mathbf{w}^{g}_{l,i})|}{dim(\mathbf{w}^{g}_{l,i})},
\end{equation}
where $m$ is the number of clients in group $g$, $\mathbf{w}_{l,i}$ represents the weights of the $l^{th}$ layer in the model for client $i$, and $\mathbf{w}^{g}_{l,i}$ represents the weights of the $l^{th}$ layer in the group model for client $i$. $D^t_{g,l}$ measures the difference between the model weights of clients and their group models for the $l^{th}$ layer in the model. 
A low $D^t_{g,l}$ means that for layer $l$, the differences of the model weights between the clients and its group model are small, and thus less frequent synchronization (i.e., aggregation) is required because the weights among clients within this group are similar.
On the other hand, a high $D^t_{g,l}$ indicates a high discrepancy for the corresponding layer, requiring frequent aggregation.
We use $\tau_{g,l}$ to denote the layer aggregation frequency, i.e., for the $l^{th}$ layer in group $g$, we aggregate it every $\tau_{g,l}$ rounds.
In our design, a layer is considered a \textit{ low-discrepancy} layer when the ratio between the discrepancy of this layer and the model discrepancy does not exceed 0.1. Otherwise, it is a \textit{high-discrepancy} layer. We use two parameters to control $\tau_{g,l}$ to distinguish the aggregation frequency for \textit{low-discrepancy} and \textit{high-discrepancy} layers: (i) for a \textit{high-discrepancy} layer, it is aggregated every $\tau$ rounds; (ii) for a \textit{low-discrepancy} layer, 
it is aggregated less frequently, i.e., every $\alpha\tau$ rounds, where $\alpha>1$. 
Suppose the total number of layers in the model is $L$, and let $D^t_{g}$ denote the average discrepancy of all model layers in group $g$. We have:
\begin{equation}\small
\setlength{\abovedisplayskip}{3pt}
\setlength{\belowdisplayskip}{3pt}
\label{eq:tau_l}
\tau_{g,l} = 
    \left\{\begin{array}{ll}
         & \alpha\tau, \quad if \quad D^t_{g,l} < 0.1*D^t_{g} \\
         &  \tau, \quad if \quad D^t_{g,l} \geq 0.1*D^t_{g}.
    \end{array}
    \right.
\end{equation}
As such, DC-PFL less frequently aggregates the layers with small discrepancies to reduce the total communication cost. It is worth noting that the whole model weights within one group are fully synchronized every $\alpha \tau$ communication rounds.

\vspace{-0.05in}
\section{Performance Evaluations}
\label{sec:evaluation}


\subsection{Evaluation Setup}
\label{sec:setup}

We evaluated the performance of different algorithms on two datasets, CIFAR-10~\cite{krizhevsky-CIFAR2009} and FashionMNIST (FMNIST)~\cite{xiao-fmnist2017}, using 50 clients. The CIFAR-10 dataset contains 50,000 color images for training and 10,000 color images for testing, with each image belonging to one of 10 classes. 
The FMNIST dataset consists of 60K training and 10K testing grayscale images of 10 different fashion items.
For both datasets, we used a convolutional neural network (CNN) with the same architecture as that in \cite{wang-infocom2020}, consisting of two convolutional layers and three fully-connected (FC) layers.

\textbf{Methods for comparison.}
We evaluate DC-PFL against the following four baselines.

1) \textbf{Standalone}: Each client forms its own group and trains the model locally using its local data.

2) \textbf{FedAvg} \cite{haddadpour-nips2019}: clients send updated local parameters to the central server after multiple epochs of local training and download the aggregated global model. This algorithm is equivalent to always training a global model using one group.

3) \textbf{Per-FedAvg} \cite{fallah-nips2020}: it finds the shared global model first, then each client adapts the global model to local data by performing a few steps of gradient descent locally.

4) \textbf{ClusterFL} \cite{ouyang-tsn2022}: this algorithm clusters clients by K-means. Each client's model is updated based on both its local data and other clients' models in the same cluster. 

\textbf{Hyperparameters.}
As the base configuration, we set the local batch size to 32 and assign every client an equally sized subset of the training data.
For each client, the data transmission rate is initialized to be 
2 Mbps by default. 
To construct a dataset for each client, we randomly 
choose a primary class ($\sigma_{p}$\% of the client's dataset) and a secondary class ($\sigma_{s}$\% of the client's dataset). 
The remaining data is uniformly distributed among the other eight classes. 
To generate heterogeneous data for different clients, we vary $\sigma_{p}$\% from 40\% to 60\% and $\sigma_{s}$\% from 20\% to 40\%. 
We calculate the model discrepancy after 5 training rounds.
For the dynamic clustering algorithm, we set $\lambda_{\tilde{\gamma}}$ to be 0.2, $t^{obv}$ to be 3, and $t^{sp}$ to be 6.
In addition, we set the base aggregation interval $\tau$ to be 5 epochs, and the interval increase factor $\alpha$ to be 3. 
For the baselines, we set their hyperparameters (as shown above) the same as those suggested in the corresponding literature.


\vspace{-0.05in}
\subsection{Evaluation Against Baselines}
\label{sec:baseline_comp}

\begin{figure}[t!]
    \subfigure[CIFAR-10]{
    \centering
        \includegraphics[width=1.63in]{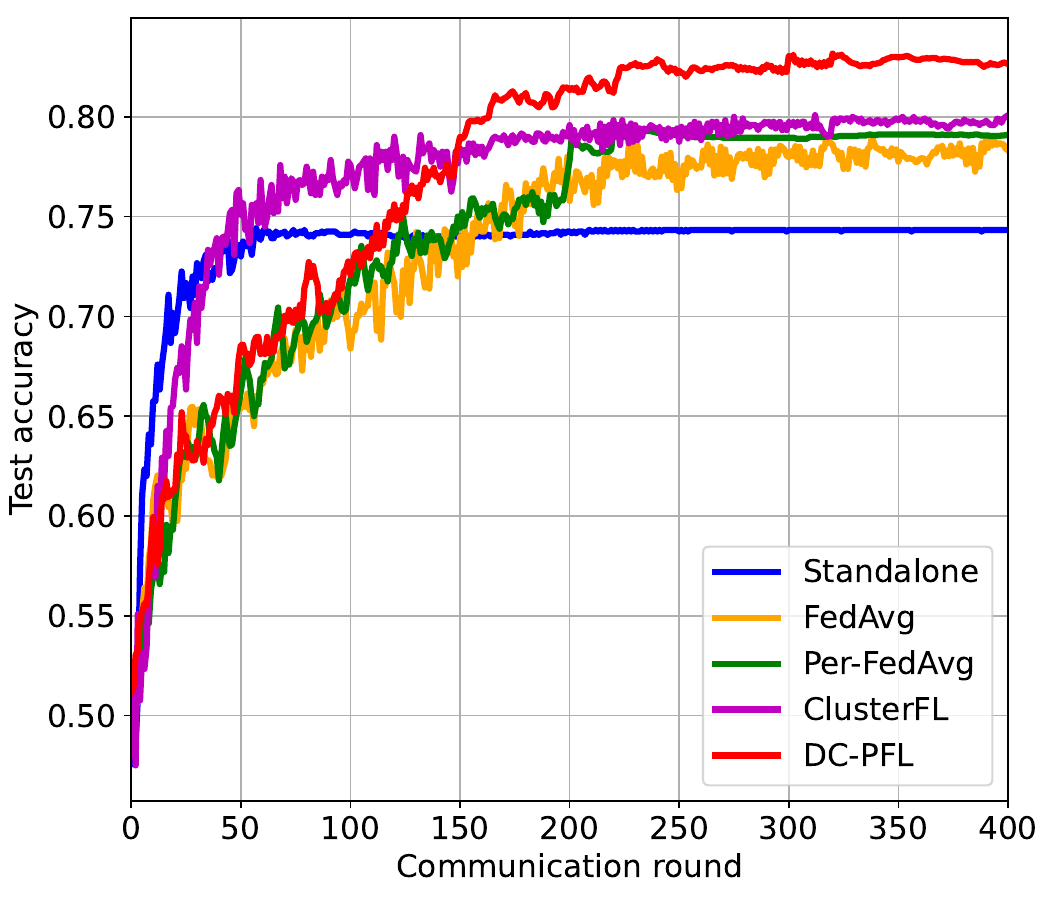}
    }
    \hspace*{-1.0em}
    \subfigure[FMNIST]{
    \centering
        \includegraphics[width=1.63in]{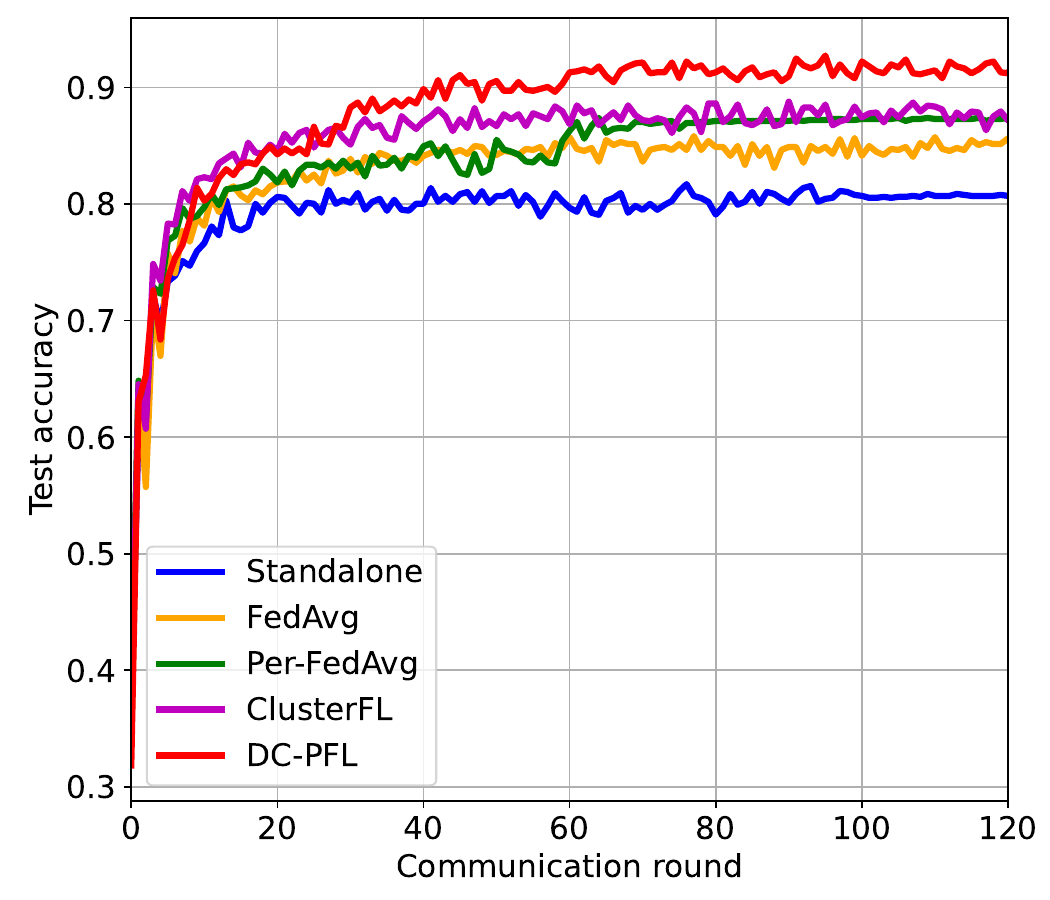}
    }
    \hspace*{-1.0em}\\
    \vspace{-0.15in}
    \caption{Accuracy v.s. Communication Round.}
\label{fig:acc_round_compare}
\vspace{-0.3in}
\end{figure}

Fig.~\ref{fig:acc_round_compare} shows the test accuracy over communication rounds on both datasets. 
We observe that DC-PFL achieves the highest final model accuracy on both datasets, with 82.7\% on CIFAR-10 and 91.3\% on FMNIST, improving the accuracy by 2.7\% and 3.4\% compared to the best-performing baseline (i.e., ClusterFL).
Among the baselines, Standalone achieves the lowest final accuracy, even though it achieves relatively high accuracy in the beginning (on the CIFAR-10 dataset). This is because Standalone only utilizes clients' local datasets for training, making it converge faster but overfit to local training datasets.
FedAvg achieves the second-lowest final accuracy because it only trains one global model for all clients, without personalizing the model for different clients. On the other hand, 
Per-FedAvg and ClusterFL achieve higher accuracy than FedAvg since they consider model personalization. However, they have their own drawbacks. Per-FedAvg first learns a global model in a global training phase and then performs local fine-tuning to personalize the model based on the local dataset. The performance of Per-FedAvg highly depends on the quality of the learned global model, which may not work well when there is high data heterogeneity, leading to poor performance on PFL. 
ClusterFL clusters clients based on the similarity of their data distributions and learns different models for different groups. However, the cluster structure is fixed throughout the training process, which may not be able to extract general features well in the early rounds. Therefore, Per-FedAvg and ClusterFL achieve only limited improvements compared to FedAvg and perform worse than DC-PFL.

\begin{figure}[t!]
    \subfigure[CIFAR-10]{
    \centering
        \includegraphics[width=1.63in]{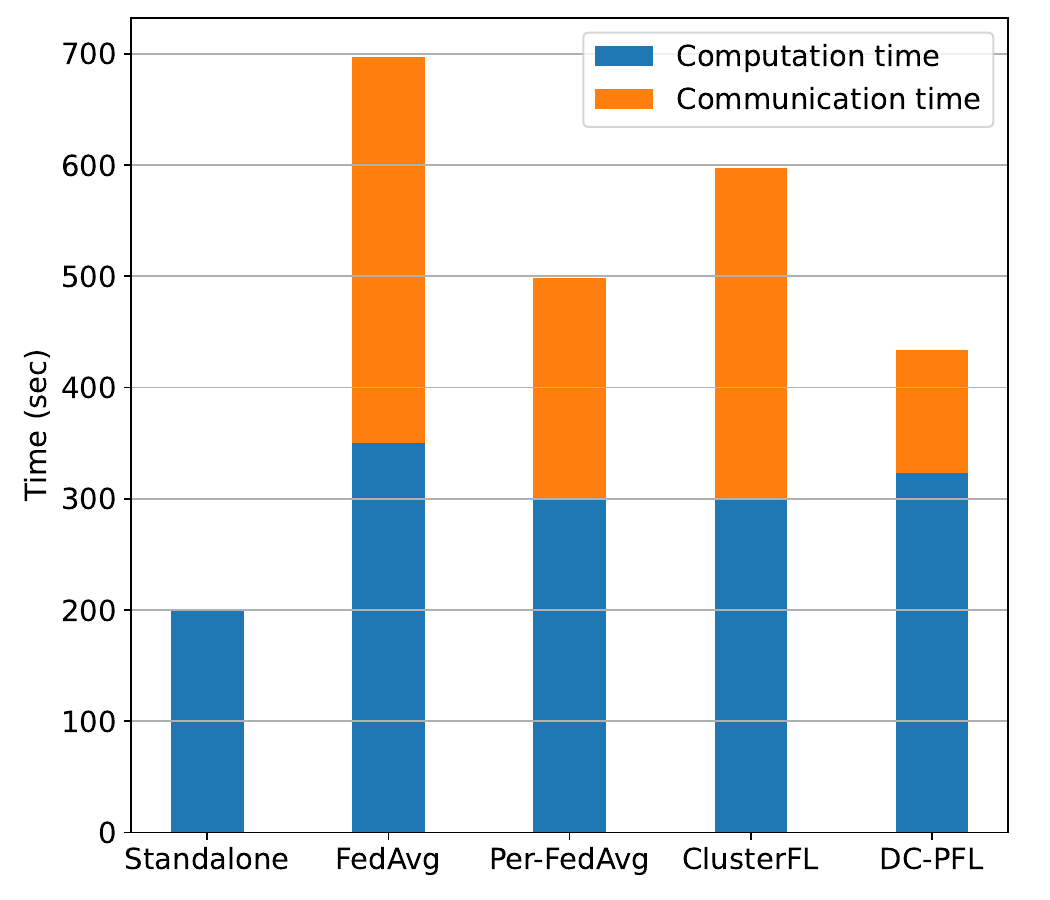}
    }
    \hspace*{-1.0em}
    \subfigure[FMNIST]{
    \centering
        \includegraphics[width=1.63in]{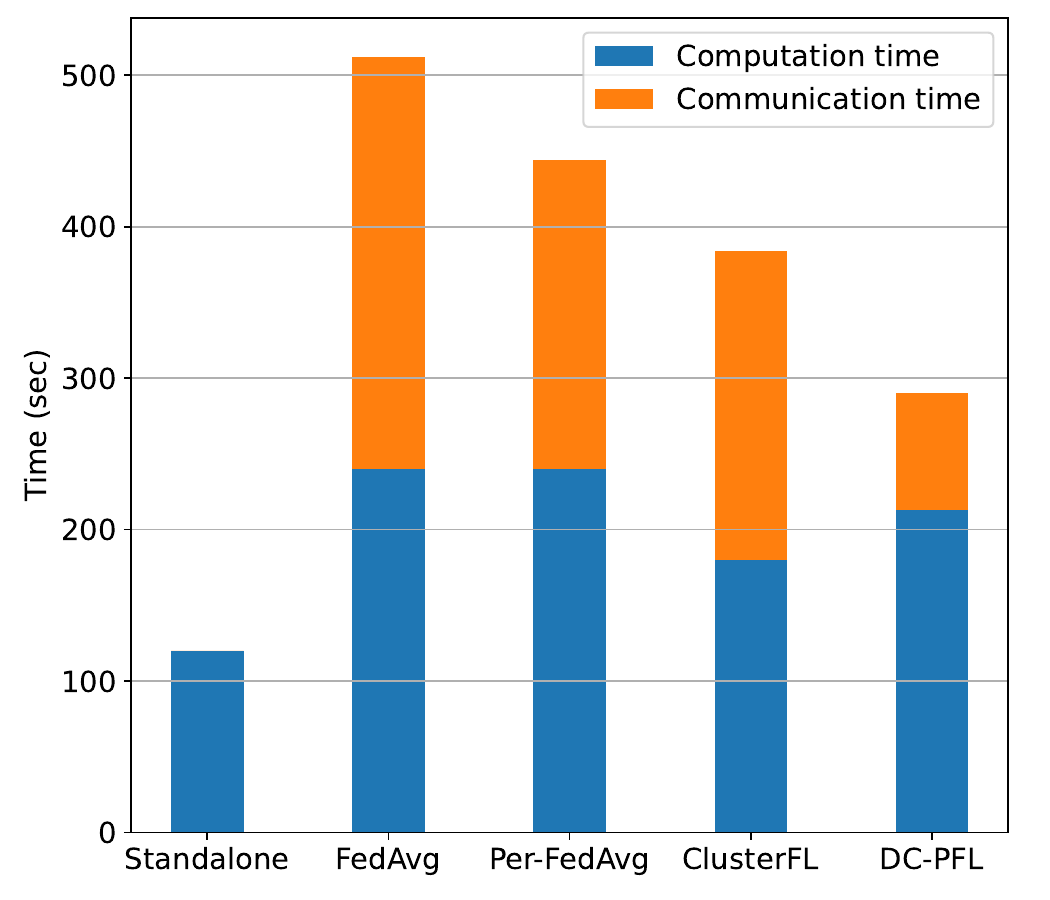}
    }
    \hspace*{-1.0em}\\
    \vspace{-0.2in}
    \caption{Comparison of total training time.}
    \label{fig:time_compare}
        \vspace{-0.25in}
\end{figure}

Fig.~\ref{fig:time_compare} shows the total training time of all algorithms when their training converges 
(in Fig.~\ref{fig:acc_round_compare}). 
Among them, Standalone has the least training time, because its training converges fast and it does not need to communicate with the server or other clients. However, the accuracy achieved by this algorithm is quite low.
FedAvg has the longest training time, mainly because 1) it has the slowest training convergenc with the most training rounds compared to other algorithms (in Fig.~\ref{fig:acc_round_compare}), 2) it transmits all model weights in every communication round, leading to longer communication time.  

Compared to FedAvg, Per-FedAvg and ClusterFL reduce the training time by 33.3\% and 14.3\% on the CIFAR-10 dataset, and reduces the training time by 13.2\% and 20.1\% on FMNIST, respectively.
Per-FedAvg spends much less communication time and thus less training time than FedAvg. This is because in Per-FedAvg, after finding an initial model, clients train their own models locally without any further communications with the server and thus reducing the communication time.
From Fig.~\ref{fig:acc_round_compare}, we observe that ClusterFL converges faster than FedAvg because it   
clusters clients with similar data into the same group and learns a model within the group. As a result,  
ClusterFL has both less computation time and less communication time than FedAvg as shown in Fig.~\ref{fig:time_compare}.

From the figure, we can see that DC-PFL spends less training time than all algorithms except Standalone. Although DC-PFL may take more rounds to converge compared with Per-FedAvg and ClusterFL, it adopts layer-wise aggregation which reduces the aggregation frequencies for certain layers, and reduces the communication time as shown in Fig.~\ref{fig:time_compare}.
For example, DC-PFL reduces the communication time by 44.2\% and 62.7\% compared to Per-FedAvg and ClusterFL on CIFAR-10, which validates the advantage of layer-wise aggregation.

\vspace{-0.05in}
\subsection{Ablation Study}
\label{sec:ablation_study}

\begin{figure}[t]
    \subfigure[CIFAR-10]{
    \centering
        \includegraphics[width=1.63in]{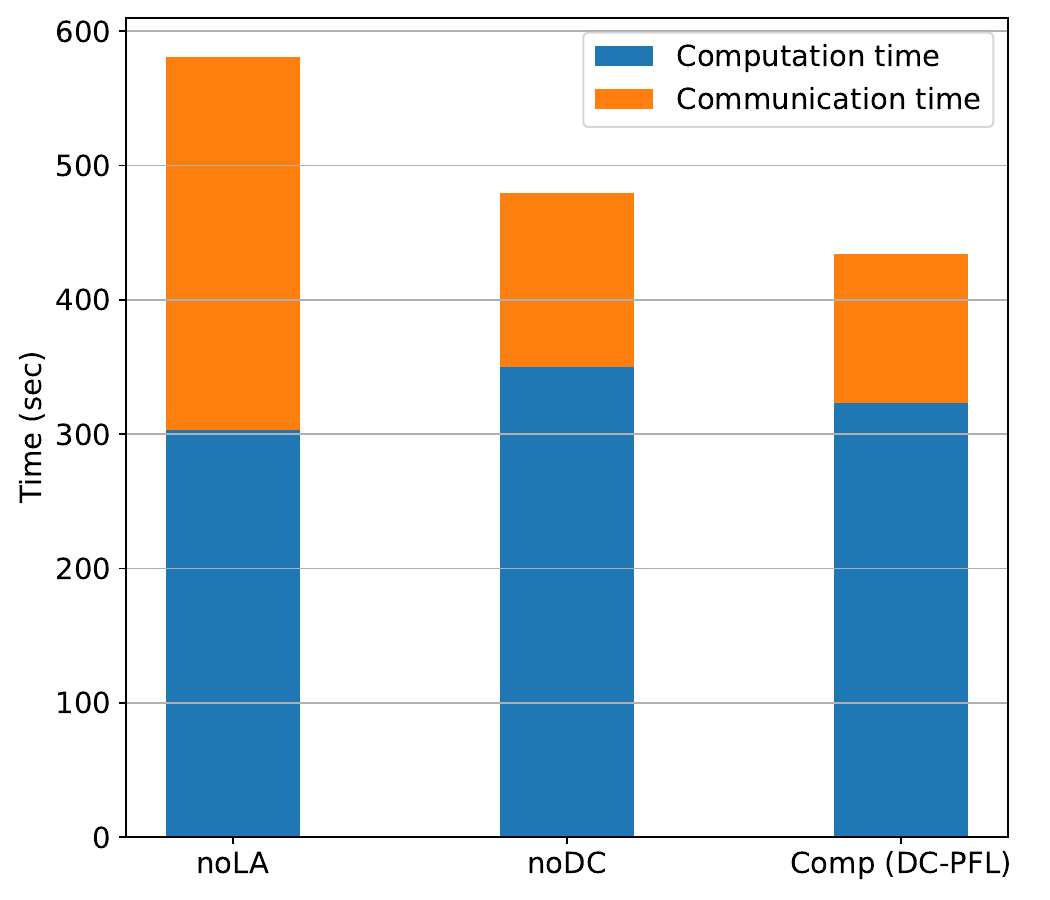}
    }
    \hspace*{-1.0em}
    \subfigure[FMNIST]{
    \centering
        \includegraphics[width=1.63in]{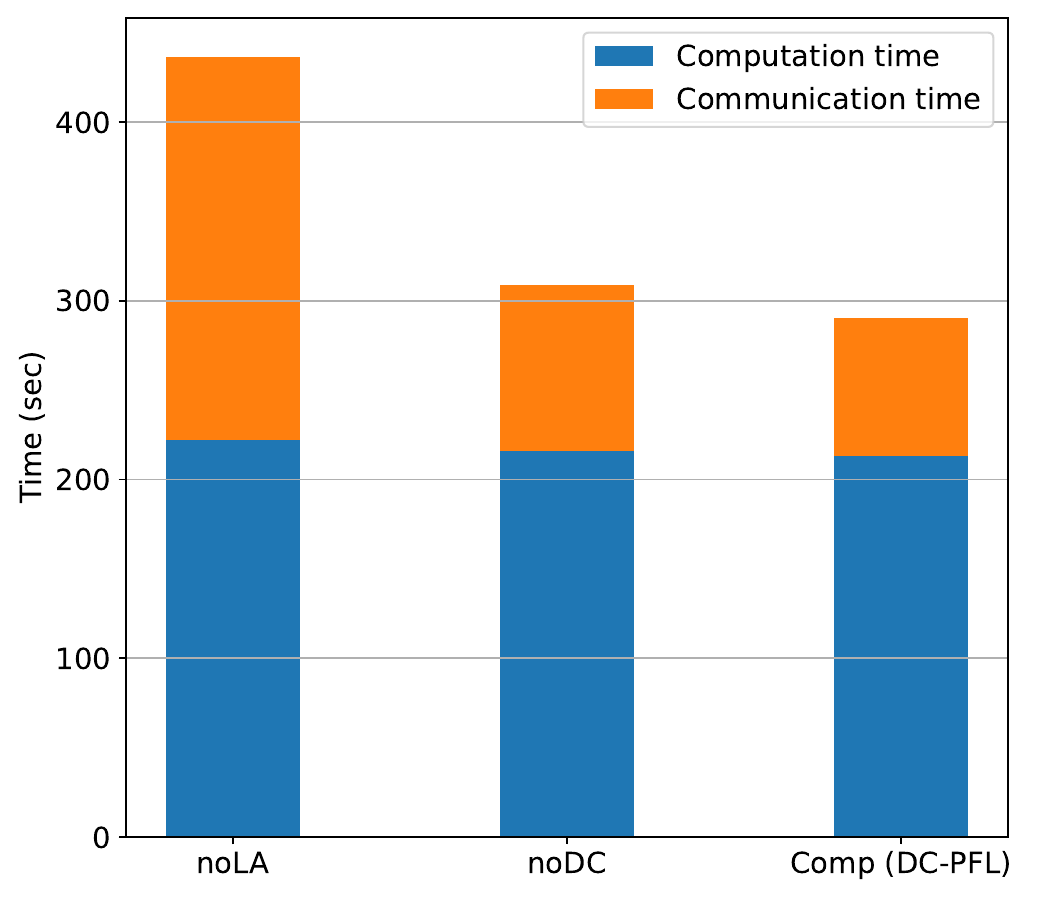}
    }
    \hspace*{-1.0em}\\
    \vspace{-0.2in}
    \caption{Ablation Study: Total Training Time.}
        \vspace{-0.25in}
    \label{fig:abl_time_compare}
\end{figure}

In this section, we conduct ablation studies to demonstrate the importance of various ideas in DC-PFL on performance. We consider three variants: 1) \texttt{Comp}, the complete version of DC-PFL with all parts included; 2) \texttt{NoLA}, a DC-PFL variant without layer-wise aggregation; 3) \texttt{NoDC}, a DC-PFL variant without dynamic clustering, where $\tilde{\gamma}$ is predetermined and the group structure is fixed. We set $\tilde{\gamma}$ to be 0.8 in \texttt{NoDC}, which achieves the highest final model accuracy when $\tilde{\gamma}$ is fixed.

\begin{figure}[htbp]
    \vspace{-0.1in}
    \subfigure[CIFAR-10]{
    \centering
        \includegraphics[width=1.63in]{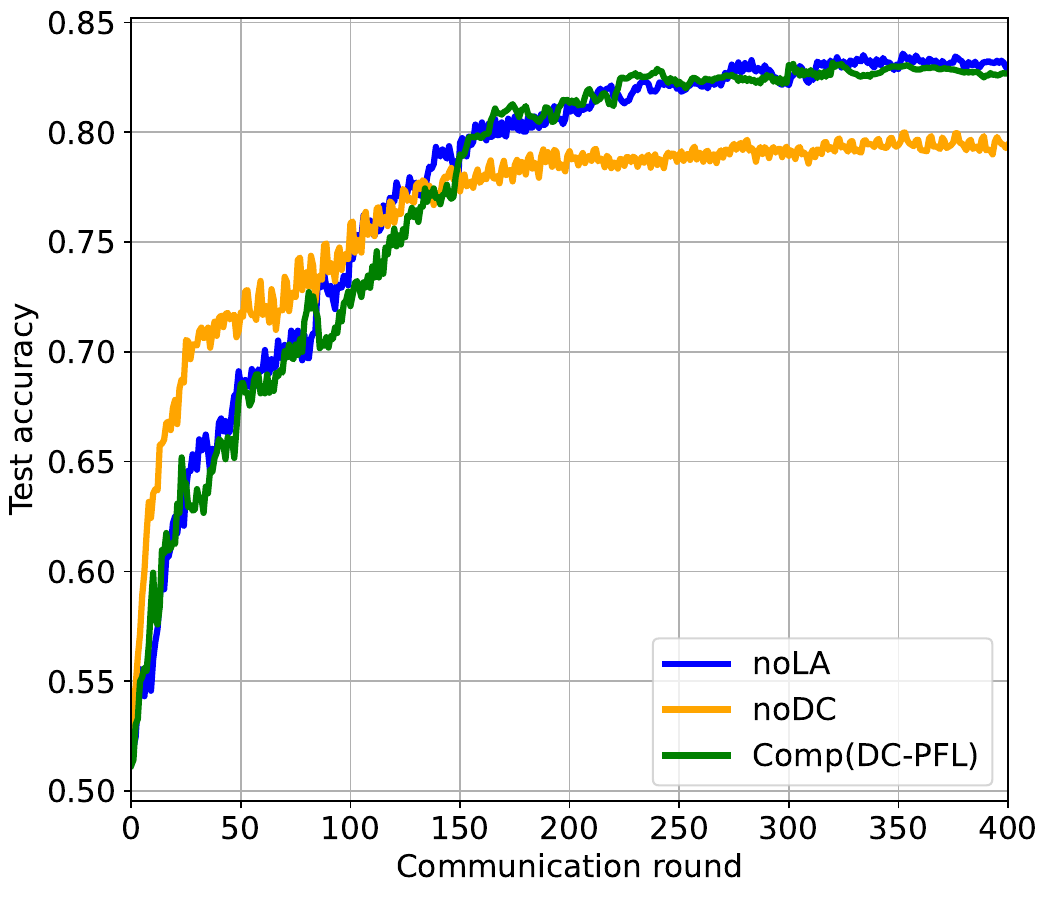}
    }
    \hspace*{-1.0em}
    \subfigure[FMNIST]{
    \centering
        \includegraphics[width=1.63in]{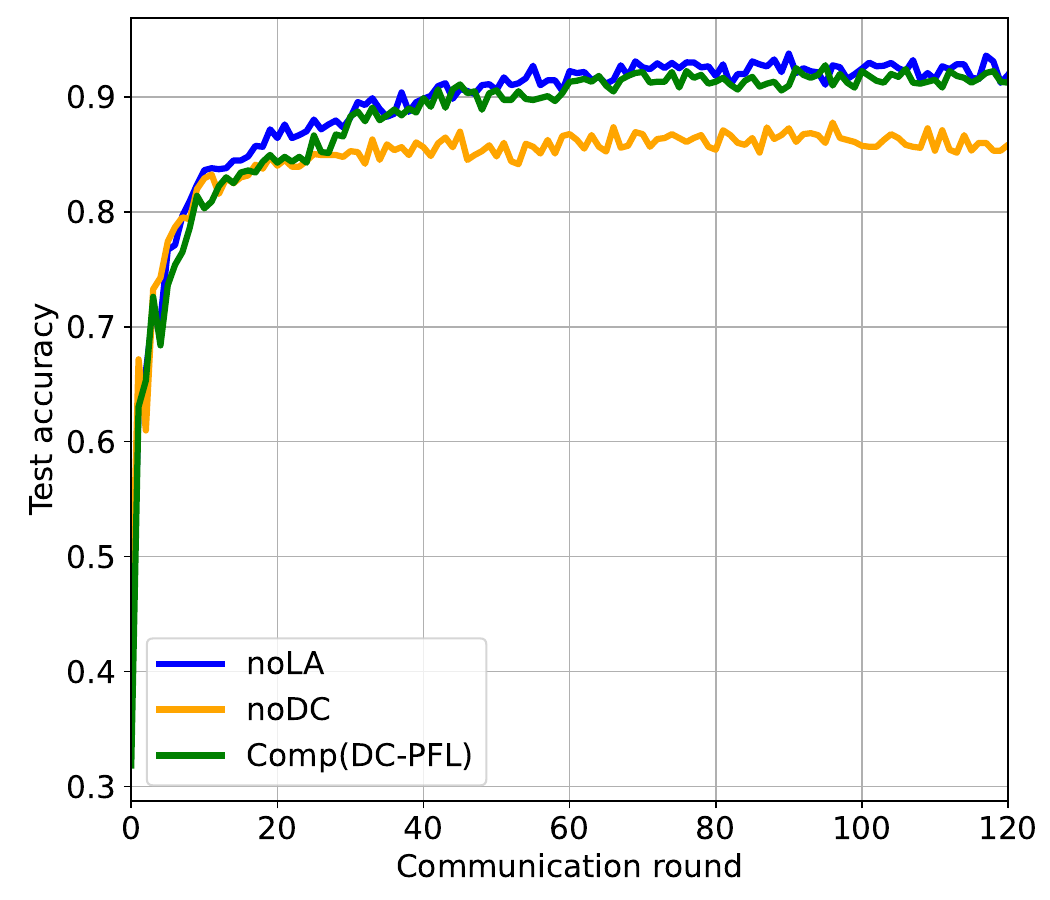}
    }
    \hspace*{-1.0em}\\
    \vspace{-0.2in}
    \caption{Ablation Study: Accuracy.}
        \vspace{-0.15in}
    \label{fig:abl_acc_compare}
\end{figure}

Fig.~\ref{fig:abl_acc_compare} shows the average accuracy change along the training time of the different variants for DC-PFL on CIFAR-10 and FMNIST datasets. Fig.~\ref{fig:abl_time_compare} shows the computation time and communication time spent by these variants until their training converges.

Compared to {\it Comp}, {\it NoDC} degrades accuracy by 3.1\% on CIFAR-10 and 6.2\% on FMNIST, while {\it NoLA} achieves similar accuracy to {\it Comp} on both datasets, with several more rounds of training.
The observation suggests that dynamic clustering plays an important role in improving model accuracy in DC-PFL. 
Compared to {\it Comp}, {\it NoLA} increases the training time by 37.1\% and 49.7\% on CIFAR-10 and FMNIST, respectively. 
From Fig.~\ref{fig:abl_time_compare}(a), we observe that the computation time of {\it Comp} is slightly higher than {\it NoLA}, which is because {\it Comp} takes several more rounds to converge. 
More important, the communication time of {\it Comp} is significantly reduced compared with {\it NoLA}, 
which clearly demonstrates the advantage of layer-wise aggregation.

\vspace{-0.05in}
\subsection{The Impact of Data Heterogeneity}
\label{sec:diff_hetero_level}

\begin{table}[t]
    \centering
    \resizebox{\columnwidth}{!}{
    \begin{tabular}{m{0.5cm}<{\centering} m{1.6cm}<{\centering} m{0.9cm}<{\centering} m{0.9cm}<{\centering} m{1.2cm}<{\centering} m{2.0cm}<{\centering} }
    \hline
    \hline
    $\bar{D}_{KL}$ & Method  & Final Acc. &  Commu. rounds & Commu. Time (s) & Total time (s)\\
    \hline
    \hline
    \multirow{6}{*}{Low}
    & Standalone  & 0.750 & -  & - & 195 ($\times$ 0.47) \\ \cline{2-6}
    & FedAvg  & 0.802 & 344  & 338 & 685 ($\times$ 1.65) \\ \cline{2-6}
    & Per-FedAvg & 0.814 & 189  & 187 & 507 ($\times$ 1.22) \\ \cline{2-6}
    & ClusterFL  & 0.817 & 286  & 276 & 569 ($\times$ 1.37) \\ \cline{2-6}
    & \textbf{DC-PFL}  & \textbf{0.843} & \textbf{288}  & \textbf{107} & \textbf{415 ($\times$ 1.00)} \\ 
    \hline
    \hline
    \multirow{6}{*}{Mid}
    & Standalone  & 0.743 & -  & - & 204 ($\times$ 0.47) \\ \cline{2-6}
    & FedAvg  & 0.784 & 352  & 331 & 697 ($\times$ 1.59) \\ \cline{2-6}
    & Per-FedAvg & 0.791 & 203  & 201 & 498 ($\times$ 1.14) \\ \cline{2-6}
    & ClusterFL  & 0.800 & 302  & 289 & 597 ($\times$ 1.37) \\ \cline{2-6}
    & \textbf{DC-PFL}  & \textbf{0.827} & \textbf{305} & \textbf{113} & \textbf{437 ($\times$ 1.00)} \\ 
        \hline
        \hline
    \multirow{6}{*}{High}
    & Standalone  & 0.741 & - & - & 220 ($\times$ 0.49) \\ \cline{2-6}
    & FedAvg  & 0.739 & 389  & 376 & 774 ($\times$ 1.71) \\ \cline{2-6}
    & Per-FedAvg & 0.761 & 253  & 251 & 581 ($\times$ 1.28) \\ \cline{2-6}
    & ClusterFL  & 0.774 & 324  & 271 & 601 ($\times$ 1.33) \\ \cline{2-6}
    & \textbf{DC-PFL}  & \textbf{0.817} & \textbf{318} & \textbf{118} & \textbf{453 ($\times$ 1.00)} \\ 
    \hline
    \hline
    \end{tabular}
    }
    \vspace{0.01in}
    \caption{Evaluation on CIFAR-10 under different $\bar{D}_{KL}$}
    \label{tab:hetero_level_CIFAR}
\vspace{-0.4in}
\end{table}

In this section, we evaluate the performance of DC-PFL against baselines under various levels of data heterogeneity (i.e., $\bar{D}_{KL}$). 
Tables~\ref{tab:hetero_level_CIFAR} and \ref{tab:hetero_level_fmnist} show the results obtained by training models on CIFAR-10 and FMNIST, respectively. 
We evaluate the algorithms based on the final model accuracy, the total number of communication rounds, the communication time (in seconds), and the total time (in seconds) when training converges (the number in the total time column after ``$\times$'' is the ratio between each baseline's total time and DC-PFL's total time). 
Note that the communication round refers to the round where clients transmit data to the server, which may be smaller than the total training round. For example, for Per-FedAvg and DC-PFL, clients may perform local model training after a certain period, which does not entail any communication rounds. 
We repeat the evaluation three times and report the average of those metrics.

From Table~\ref{tab:hetero_level_CIFAR}, we observe that DC-PFL outperforms all baseline algorithms in terms of accuracy by up to 9.3\% under various data heterogeneity. 
Here, we repeat the data generation process 100 times and record the data heterogeneity in each iteration. 
We then divide the range between the lowest and highest heterogeneity values into three equal-length sub-intervals labeled as ``Low'', ``Mid'', and ``High'' respectively. This enables us to categorize and label the data heterogeneity using these three intervals.
From the table, we can also see that DC-PFL has the lowest total training time among 
all algorithms except Standalone (as shown by the ratios).
Among the baselines, Standalone does not need any communication and it has the least total training time. 
However, it has much lower accuracy due to the lack of training data, e.g., the lowest accuracy when $\bar{D}_{KL}$ is ``Low'' (0.750) and ``Mid'' (0.743), and the second lowest accuracy when the data heterogeneity is ``High'' (0.741).
When the data heterogeneity is ``High'' (0.739), FedAvg has the lowest accuracy, which supports our hypothesis that a global model cannot adapt to different clients' data simultaneously under high data heterogeneity. 

From the two tables, we observe that 
the level of data heterogeneity affects the performance of FedAvg, Per-FedAvg, ClusterFL, and DC-PFL: a higher level of data heterogeneity in general degrades the final model accuracy of these algorithms. 
Among them, DC-PFL achieves the highest accuracy under all levels of $\bar{D}_{KL}$. 
This is because DC-PFL reduces the data heterogeneity by clustering clients with similar data distributions and avoids overfitting by starting with a large cluster and splitting the cluster into smaller ones gradually.    
A higher level of data heterogeneity also decreases the convergence speed of all algorithms, 
which results in more communication rounds and thus longer total training time. 
However, DC-PFL has the least total time under various $\bar{D}_{KL}$ due to the 
advantage of layer-wise aggregation, which greatly reduces the communication time in each round.
Finally, the evaluation results on the FMNIST dataset are consistent with those on the CIFAR-10 dataset, indicating the robustness of DC-PFL across different datasets.

\begin{table}[t]
    \centering
    \resizebox{\columnwidth}{!}{
    \begin{tabular}{m{0.5cm}<{\centering} m{1.6cm}<{\centering} m{0.9cm}<{\centering} m{0.9cm}<{\centering} m{1.2cm}<{\centering} m{2.0cm}<{\centering} }
    \hline
    \hline
    $\bar{D}_{KL}$ & Method  & Final Acc. &  Commu. rounds & Commu. time (s) & Total time (s)\\
    \hline
    \hline
    \multirow{6}{*}{Low}
    & Standalone  & 0.810 & - & -  & 114 ($\times$ 0.43) \\ \cline{2-6}
    & FedAvg  & 0.876 & 79  & 268 & 506 ($\times$ 1.91) \\ \cline{2-6}
    & Per-FedAvg & 0.890 & 58 & 197 & 431 ($\times$ 1.63) \\ \cline{2-6}
    & ClusterFL  & 0.901 & 61 & 207 & 391 ($\times$ 1.48) \\ \cline{2-6}
    & \textbf{DC-PFL}  & \textbf{0.925} & \textbf{57} & \textbf{73} & \textbf{265 ($\times$ 1.00)} \\ 
    \hline
    \hline
    \multirow{6}{*}{Mid}
    & Standalone  & 0.808 & - & - & 120 ($\times$ 0.41) \\ \cline{2-6}
    & FedAvg  & 0.852 & 82 & 262 & 512 ($\times$ 1.77) \\ \cline{2-6}
    & Per-FedAvg & 0.873 & 61 & 207 & 444 ($\times$ 1.53) \\ \cline{2-6}
    & ClusterFL  & 0.879 & 64 & 214 & 409 ($\times$ 1.41) \\ \cline{2-6}
    & \textbf{DC-PFL}  & \textbf{0.913} & \textbf{60} & \textbf{78} & \textbf{290 ($\times$ 1.00)} \\ 
        \hline
        \hline
    \multirow{6}{*}{High}
    & Standalone  & 0.802 & - & - & 119 ($\times$ 0.38)  \\ \cline{2-6}
    & FedAvg  & 0.801 & 91 & 309 & 582 ($\times$ 1.88) \\ \cline{2-6}
    & Per-FedAvg & 0.845 & 72 & 245 & 501 ($\times$ 1.62) \\ \cline{2-6}
    & ClusterFL  & 0.852 & 70 & 233 & 448 ($\times$ 1.45) \\ \cline{2-6}
    & \textbf{DC-PFL}  & \textbf{0.896} & \textbf{64} & \textbf{82} & \textbf{310 ($\times$ 1.00)} \\ 
    \hline
    \hline
    \end{tabular}
    }
    \vspace{0.01in}
    \caption{Evaluation on FMNIST under different $\bar{D}_{KL}$}
    \label{tab:hetero_level_fmnist}
    \vspace{-0.4in}
\end{table}

\vspace{-0.05in}
\section{Related Work}
\label{sec:rel_work}

\textbf{Personalized Federated Learning.}
PFL addresses data heterogeneity by adapting a global model to individual client data. Existing PFL methods are broadly classified into data-based \cite{fallah-nips2020, hanzely-arxiv2020federated, smith-nips2017, ghosh-nips2020, ouyang-tsn2022, Tu-sensys2021} and model-based \cite{arivazhagan-arxiv2019, li-mobicom2021, li-2021sensys, huang-aaai2021} approaches. Data-based methods tackle heterogeneity by leveraging data statistics without altering model architectures, while model-based approaches customize the model structure or parameters for each client.


Data-based PFL often employs local fine-tuning, where a globally learned model is adapted to each client's local data. Examples include Per-FedAvg \cite{fallah-nips2020} using second-order optimization, a mixture learning approach \cite{hanzely-arxiv2020federated}, Multi-task learning (MOCHA) \cite{smith-nips2017}, and clustering similar clients for collaborative learning \cite{ouyang-tsn2022}.
Model-based PFL personalizes models by modifying their structure or parameters. This includes methods where clients share base layers but have unique personalization layers \cite{arivazhagan-arxiv2019}, clients learn personalized sparse sub-networks \cite{li-mobicom2021}, and approaches using federated attentive message passing to enhance collaboration among similar clients \cite{huang-aaai2021}. 
Wang et al. \cite{pFedHR} proposed pFedHR which leverages heterogeneous model reassembly to generate optimal personalized model structures for different clients.
Chen et al. \cite{FedRoD} proposed FedRoD as a unifying framework that allows the FL system to achieve high generic and personalized performance simultaneously by formulating different personalized adaptive predictors that minimize the empirical risks of local clients.
Zhou et al. \cite{zhou2024eafl} proposed EAFL which dynamically clusters clients with similar gradient directions to mitigate the straggler effect on the accuracy of the global model.

We note that our work falls into the category of data-based PFL approaches and is orthogonal to model-based approaches\cite{arivazhagan-arxiv2019, li-mobicom2021, FedRoD}. In addition, \cite{zhou2024eafl} focuses on generic FL, while \cite{pFedHR} assumes different model structures for clients. Unlike existing work, We analyze the influence of data heterogeneity and data amount in different training rounds and propose a dynamic clustering algorithm for PFL.

\textbf{Communication-Efficient Federated Learning.}
Communication cost reduction is an important issue in FL. Existing methods include gradient compression, which reduces the data transmitted per round through sparsification or quantization \cite{ozfatura-isit2021, liu-infocom2023, guan2024fedtc}, and periodic aggregation, where clients perform multiple local updates before less frequent uploads \cite{zhao-icdcs2019, haddadpour-nips2019, reisizadeh-aistats2020}. 
For example, dynamic staleness thresholds \cite{zhao-icdcs2019} and combining periodic averaging with quantization \cite{reisizadeh-aistats2020} have recently been proposed. However, these methods often apply a uniform aggregation frequency across all layers, potentially impacting accuracy and model convergence. 
We propose a method to determine layer-specific aggregation frequencies based on inter-client discrepancies at each layer. While layer-wise aggregation has been explored \cite{lee-2023aaai}, our DC-PFL algorithm uniquely integrates dynamic clustering, enhancing the effectiveness of layer-wise aggregation by promoting similarity among clients within the same group.

\vspace{-0.05in}
\section{Conclusions}
\label{sec:conclusion}

In this paper, we proposed a dynamic clustering algorithm, DC-PFL, for personalized federated learning to address the issue of data heterogeneity among the clients.
Our approach leverages model discrepancy to estimate data heterogeneity while ensuring data privacy. 
By dynamically clustering similar clients with varying dissimilarity thresholds during training, we aim to train models to learn general features by exploring larger groups in early training phases, and obtain personalized models for different clients by exploring smaller groups in later training phases to improve the final model accuracy.
We further aggregate different layers with different frequencies based on layer-wise discrepancies to reduce the communication costs without compromising the model accuracy.
Our evaluation results demonstrate that DC-PFL effectively improves model accuracy and reduces total training time for PFL.


\bibliographystyle{IEEEtran}
\bibliography{Heting}

\vfill

\end{document}